\documentclass[times]{article}
\usepackage{iclr2019_conference,times}

\usepackage{moreverb,url}

\usepackage[colorlinks,bookmarksopen,bookmarksnumbered,citecolor=red,urlcolor=red]{hyperref}

\usepackage{amssymb}
\usepackage{amsthm}
\usepackage{float}
\usepackage[ruled]{algorithm2e}
\usepackage{graphicx}
\usepackage{booktabs}

\usepackage{caption}
\usepackage{subcaption}
\usepackage{anyfontsize}
\usepackage{url}
\usepackage{amsfonts}
\usepackage{wrapfig}
\usepackage[toc,page,header]{appendix}
\usepackage{minitoc}

\DontPrintSemicolon
\usepackage[english]{babel}
\usepackage{natbib}
\makeatletter
\let\saved@includegraphics\includegraphics
\AtBeginDocument{\let\includegraphics\saved@includegraphics}
\renewenvironment*{figure}{\@float{figure}}{\end@float}
\makeatother

\makeatletter
\newcommand{\removelatexerror}{\let\@latex@error\@gobble}
\makeatother

\preiclrcopy
\iclrfinalcopy

\begin{document}

\title{Reverse Experience Replay}

\author{
\fontsize{8}{10}\hspace{-3px}Rotinov Egor\\
\fontsize{8}{10}Bauman Moscow State Technical University, Russia\\
\fontsize{8}{10}\texttt{rotinov@pm.me}
}

\maketitle

\begin{abstract}
This paper describes an improvement in Deep Q-learning called Reverse Experience Replay (also RER) that solves the problem of sparse rewards and helps to deal with reward maximizing tasks by sampling transitions successively in reverse order. On tasks with enough experience for training and enough Experience Replay memory capacity, Deep Q-learning Network with Reverse Experience Replay shows competitive results against both Double DQN, with a standard Experience Replay, and vanilla DQN. Also, RER achieves significantly increased results in tasks with a lack of experience and Replay memory capacity.
\end{abstract}

\section{Introduction}
Reinforcement learning with sparse rewards is a recent problem that is partially solved by \textit{Experience Replay} \citep{Lin2014} where transitions are stored for some time and can be reused for agent update more than once. However, in a standard Experience Replay, transitions for learning are sampled uniformly at random. Therefore, this type of sampling can negatively impact the agent (Q-values of some states can be entirely wrong for a long time and used for updating Q-values of other states).

This paper covers Reverse Experience Replay, one of the methods that can be used for faster learning (less experience is needed). The key idea is that the agent learns from the state in which the reward was reached all the way to the initial state (firstly, it updates the Q-value of the state one step before the reward is achieved, then the process is repeated for each consecutive step throughout the sequence of visited states). So, the Q-value of the beginning state and action tends to true Q-value. This approach is based on the \textit{Prioritized Sweeping} method \citep{Moore1993}, which was adapted for use with Neural networks. The approach is similar to the idea of \textit{Generalized Prioritized Sweeping} paper \citep{Andre1998} but can be used for model-free Reinforcement Learning. 

It is worth to mention another research with another type of Experience Replay proposed: \textit{Reward Backpropagation Prioritized Experience Replay} \citep{Zhong2017}. The key idea of papers is similar, but in there, some transitions can be thrown away without learning from them as in standard Experience Replay is. Also, there is a problem with close-in time transitions with nonzero rewards, because there is a probability of sampling them in the one minibatch that will provide noise in Q-values. An explanation of dealing with these problems is provided in the experiment section.

The importance of playing transitions in reverse order has also been proven in neuroscience research papers such as \textit{Reverse Replay of Hippocampal Place Cells Is Uniquely Modulated by Changing Reward} \citep{Ambrose2016}. In addition, the reverse order update is helpful for reward maximization tasks where the reward is granted in each transition.

\section{Background}

The discounted future reward of step $t$ with the discount factor $\gamma$ for trajectory $\tau$ can be represented as follows in equation~\ref{eq:offtop}.

\begin{equation}
R_t(\tau) \equiv  \sum_{k=0}^{\infty}{\gamma_t^{(k)} \cdot r_{t+k+1}}, \ where \ \gamma \in [0, 1].
\label{eq:offtop}
\end{equation}

The goal of Q-learning is to predict the maximum of the possible discounted future reward for the state $s$ and action $a$ taken in this state. The Q-value of pre-terminate state {$s_{t-1}$} and action $a$ in a deterministic environment can be represented from the Bellman equation as follows in equation~\ref{eq:1}.

\begin{equation}
Q^{\pi}(s_{t-1}, a) =  r(s_t|s_{t-1}, a), \ where \ s_t \ is \ the \ terminate \ state.
\label{eq:1}
\end{equation}

Of other states in equation \ref{eq:2}.

\begin{equation}
\label{eq:2}
Q^{\pi}(s_i, a_i) =  r(s_{i+1}|s_i, a_i) + \gamma \cdot \max_{a'} Q^{\pi}\mbox(s_{i+1}, a').
\end{equation}

\begin{figure}[H]
\setlength{\fboxsep}{0pt}%
\setlength{\fboxrule}{0pt}%
\centering
\scalebox{0.8}[0.8]{
\includegraphics{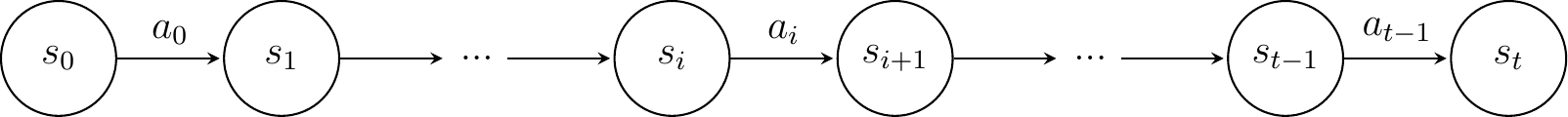}
}
\caption{Chain of transitions (Markov chain).}
\label{chain0}
\end{figure}

And all transitions in the episode can be represented as a chain. See Figure~\ref{chain0}.

\section{Reverse Experience Replay}

\subsection{Motivating example}
The problem of sparse rewards can be described in a gridworld environment where rewards for all states are zero, except two terminate states where the reward is 1 and -1. An agent sees only his position and has four possible actions: move right, left, up or down. By using Reverse Experience Replay, the agent tends multiple Q values to the true values in $N$ samples, where $N$ is the number of visited states, while if standard Experience Replay is used more samples are required on average.

\subsection{Reverse order update}

As a consequence of equations~\ref{eq:1} and~\ref{eq:2} and Figure~\ref{chain0}, the Q-value of the current state and action depends on the $\max$ Q-value of the next state in a deterministic environment and partially depends on the $\max$ Q-value in a stochastic environment. As a result, using the reverse order update of Q-values is more efficient than the forward one. 

In the case of Neural networks, it is not required to calculate the probability of transition from the current state $s$ to the next state $s'$ by action $a$ as in \textit{Generalized Prioritized Sweeping}. Transition with the biggest probability will appear more often than others at infinity. For this reason, the Q-value will be updated from this transition more times than from others, and if the learning rate is low enough, the Q-value of the state $s$ and the action $a$ will tend to the expected value of $Q(s, a)$. 

\subsection{Reverse Experience Replay}

Transitions ($s, a, r, s'$) are inserted at the beginning of the Reverse Experience Replay memory as they occur. The algorithm of sampling can be presented as follows, shown in Algorithm \ref{algo:0}.

\begin{figure}[H]
\SetAlgoNoLine
\setlength{\fboxsep}{0pt}%
\setlength{\fboxrule}{0pt}%
\centering
\removelatexerror

	\begin{algorithm}[H]
	\KwIn{$memory$, $bias$, $stride$, $batchsize$}
	\KwOut{$batch$, $bias$}
	\SetKwFunction{FMain}{Sample}
	\SetKwProg{Fn}{Function}{:}{}
	\Fn{\FMain{$memory$, $bias$, $stride$, $batchsize$}}{
		\;
		\For{$i \gets 0$ \textbf{to} $batchsize$} {
			$batch[i] \gets memory[bias + i \cdot stride]$\;
		}
		\;
		\uIf{$(bias + 2) < 2 \cdot stride$} {
			$bias \gets bias + 2$\;
		}
		\Else{
			$bias \gets 0$\;
		}
		\;
		\Return{$batch$, $bias$}
	}
	\caption{Algorithm of sampling mini-batch from the memory}
	\label{algo:0}
	\end{algorithm}
\end{figure}

\begin{figure}[H]
\centering
\scalebox{0.8}[0.8]{
\includegraphics{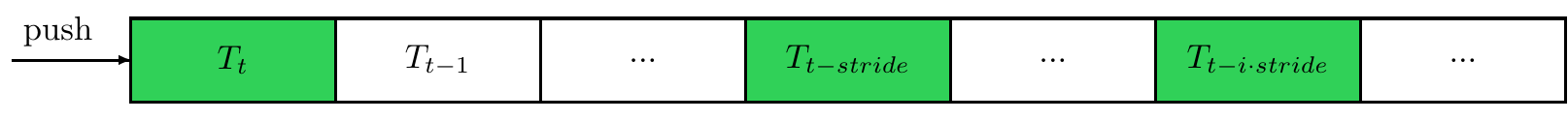}
}

\caption{Mini-batch sampling (step 1), where $T_i$ is the transition ($s_i, a_i, r(s_{i+1}|s_i, a_i), s_{i+1}$),  and highlighted transitions are chosen for the mini-batch.} 
\label{fig:sample1} 
\end{figure}

\begin{figure}[H]
\centering
\scalebox{0.8}[0.8]{
\includegraphics{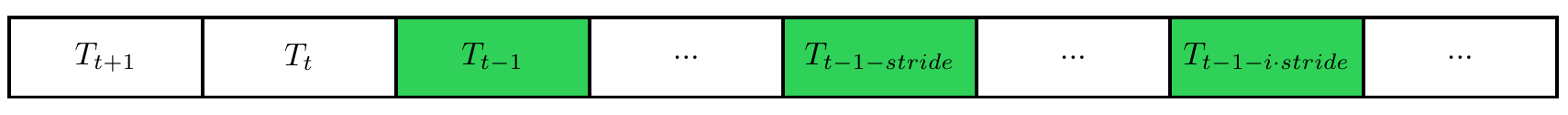}
}
\caption{Mini-batch sampling (step 2), where $T_i$ is the transition ($s_i, a_i, r(s_{i+1}|s_i, a_i), s_{i+1}$),  and highlighted transitions are chosen for the mini-batch.} 
\label{fig:sample2} 
\end{figure}

By this algorithm, Q-values from the part of the chain from Figure \ref{chain0} are being updated in reverse order. However, no computations of probabilities are needed because of the learning rate, and it is not needed to store all of the transitions (${s_i, a_i, r(s_{i+1}|s_i, a_i), s_{i+1}}$) from the chain. Figures \ref{fig:sample1} and \ref{fig:sample2} show the first and second steps of sampling by the presented algorithm respectively. In the case of limited storage capacity, Q-values are being updated starting from the different values as represented in Figure \ref{chain:1} below.

\begin{figure}[H]
\centering
\scalebox{0.85}[0.85]{
\includegraphics{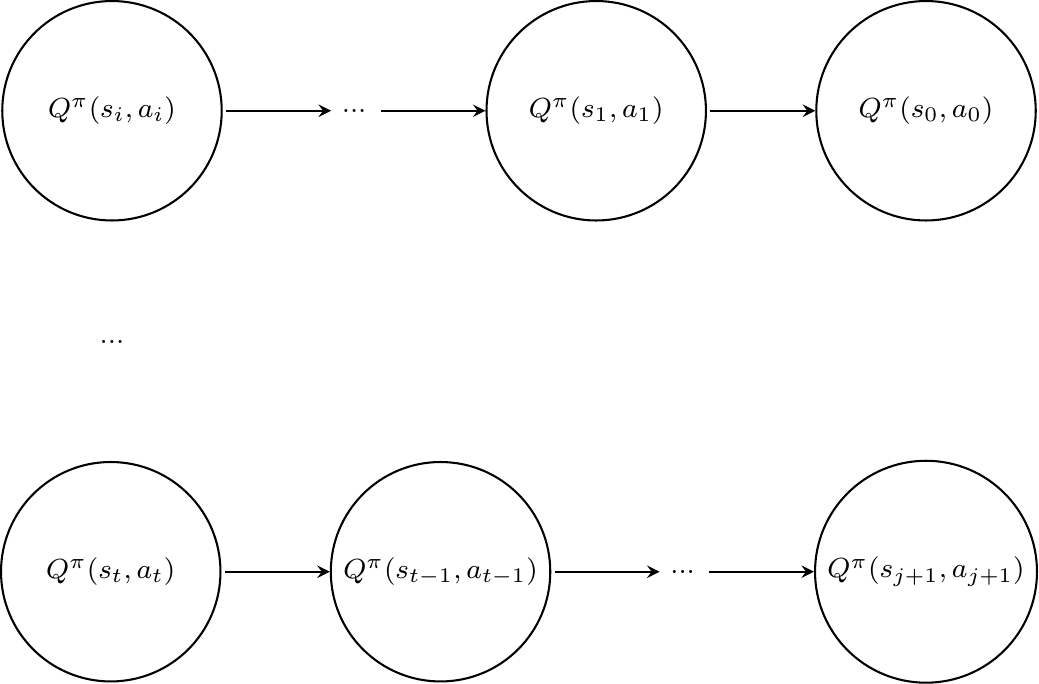}
}
\caption{Chain of Q-values updates (not all of the dependencies are direct because of the $\max$ operator).}
\label{chain:1}
\end{figure}

\section{Experiments}

The presented sampling algorithm is not stable for Neural network on its own and should be supplemented by uniformly at random sampling for nearly half of the mini-batch for revisiting old transitions. In these experiments the rest of the minibatch was sampled from the part of the RER witch will not be affected by this and future samples by Algorithm \ref{algo:0}.

RER stride should not be a multiple of the average length of the episode to avoid sampling similar but not equal states in one mini-batch. This is needed to provide less noisy Q-values for the next update.

For all experiments with RER, DQN without Target-Network was used since DQN with Target-Network will cancel out reverse order update effect because of updating from old max Q-value from Target-Network. Also, that is why \textit{Double DQN} \citep{VanHasselt2015} was not used with RER.

\subsection{Mountain Car Problem}

Mountain Car Problem is described in the book called \textit{Reinforcement Learning: An Introduction} \citep{Sutton2018}. The goal of this environment is to drive up on the mountain. However, the car's engine is not strong enough to simply accelerate and scale the mountain. Every frame agent receives -1 reward. Therefore, the dependencies of Q-values are strong. Considering these conditions, the reverse order update is useful here.

All results are the average of 3 learning and test iterations.

\begin{figure}[H]
\centering
\scalebox{0.85}[0.85]{
\includegraphics{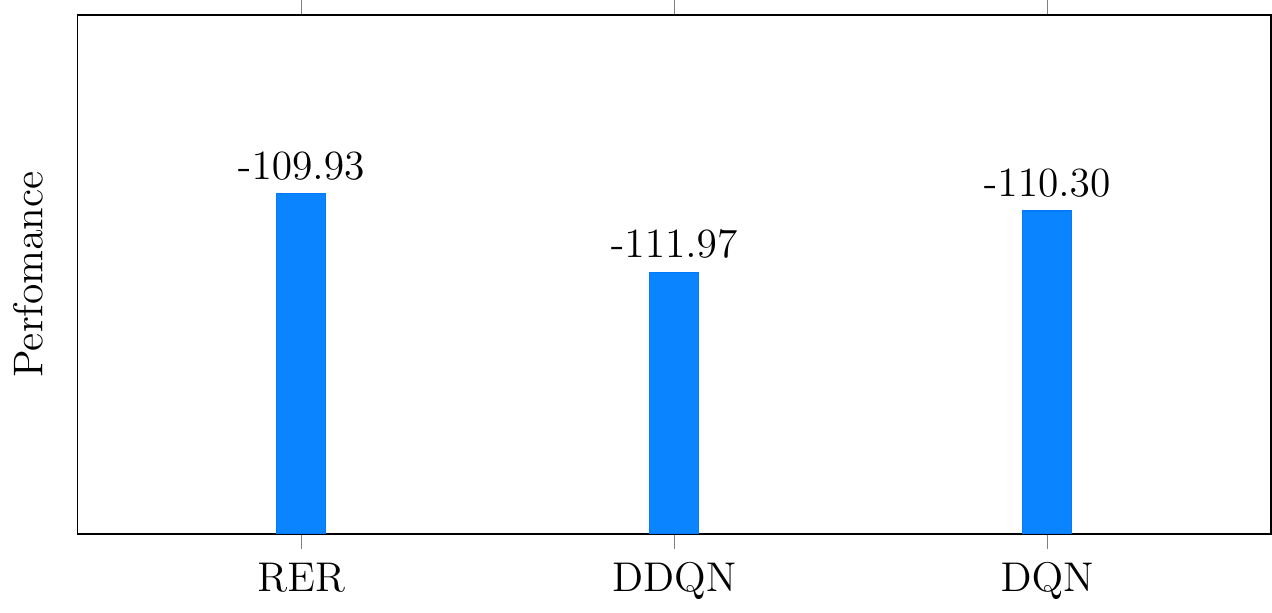}
}

\caption{Performance of DQN+RER, DDQN+ER, DQN+ER algorithms in the Mountain Car Problem (the mean of the test results of 3 different learning processes from 3 different seeds).}
\label{perfomanceMC}
\end{figure}

Deep Q-Learning Network with Reverse Experience Replay shows competitive results against Double DQN with Experience Replay and vanilla DQN with Experience Replay (Figure \ref{perfomanceMC}). Double DQN achieves the smallest results because of the Target-Network update (some transitions were sampled before Target-Network update, and the old max Q-value was used).

Table \ref{tabl1} presents the details of the Mountain Car experiment (NN structure, training and testing hyperparameters).

\begin{table}[H]
\small\sf\centering
\caption{Mountain Car experiment details}
\label{tabl1}

\begin{tabular}{llll}
\toprule
Algorithm & RER & DDQN & DQN \\
\midrule
Learning rate & 0.0035 & 0.0025 & 0.0025 \\
RER stride & 300 & --- & --- \\
Target-Network update frequency & --- & 100 & --- \\
\bottomrule
\end{tabular}\\[10pt]

\begin{tabular}{ll}
\toprule
\multicolumn{2}{c}{All algorithms} \\
\midrule
$\gamma$-discount factor & 0.99999 \\
$\epsilon$-greedy & from 1 to 0.1 \\
Decay in first n frames & 50000 \\
Memory size  & 50000 \\
Batch size & 32 \\
Final test episodes & 1000 \\
Number of hidden layers & 1 \\
Hidden layer size & 64 \\
Activation & Tanh \\
Loss function & Mean Squared Error \\
Optimizer                   & RMSprop \\ 
Oprimizer $\beta$   & 0.99 \\
Training length     & 10000 episodes \\
Env's gravity       & 0.0025 \\
Env's force of the car's engine              & 0.0025 \\
\bottomrule
\end{tabular}
\end{table}

\subsection{Atari Breakout}

Nowadays, Atari video games such as Breakout are the benchmark and many researchers use them in their significant work \cite[e.g.][]{Wang2015}. For this experiment, Dueling DQN with RER and Dueling Double DQN was used. The results of Dueling DQN with RER are better than the ones of vanilla DQN, from \textit{Human-level control through deep reinforcement learning} research paper \citep{Antonoglou2015} even with 10x lesser Experience Replay memory capacity and thus, vanilla DQN results are not presented here.

\begin{table}[H]
\small\sf\centering
\caption{Atari Breakout experiment details}
\label{tabl2}

\begin{tabular}{lll}
\toprule
Algorithm                       &         RER        & DDQN \\
\midrule
RER stride                      &    300     & ---                \\
Target-Network update frequency &       ---             & 10000            \\
\bottomrule
\end{tabular}\\[10pt]

\begin{tabular}{ll}
\toprule
\multicolumn{2}{c}{All algorithms} \\
\midrule
Learning rate                                     & 0.00003           \\
$\gamma$-discount factor          & 0.99               \\
$\epsilon$-greedy                           & from 1 to 0.2    \\
Decay in first n frames                    & 150000             \\
Memory size                                      & 100,000 transitions             \\
Batch size                                           & 32                 \\
Intermediate test episodes           & 100                \\
Intermediate test                              & every 200 episodes                \\
Final test episodes for the best    & 1000                \\
Activation                                            & ReLU               \\
Loss function                                     & Mean Squared Error \\
Optimizer                                            & RMSprop \\
Optimizer $\beta$                            & 0.99 \\
Action repetitions                             & 4 \\
Replay period                                     & every 4 agent steps \\
Replay start size                                & None \\
No-op max                                          & 30 \\
Training length                                   & 40M frames \\
\bottomrule
\end{tabular}
\end{table}

Images preprocessing, stuck frames parameters, Neural network architecture, and other parameters not mentioned above are used according to \textit{Rainbow} research paper \citep{Hessel2017}. The differences in the parameters are presented in Table \ref{tabl2}. The average length of the episode is 878.5 frames during the training and every score point matters, and that is why it is called learning from sparse rewards. Experience Replay's memory capacity is 100,000 transitions versus 1,000,000 in \textit{Rainbow} paper. There also are fewer training frames, therefore, the results are smaller as well. Dueling approach is used to show that Reverse Experience Replay can be used with this type of Neural network's architecture.

\begin{figure}[H]
\centering

\includegraphics{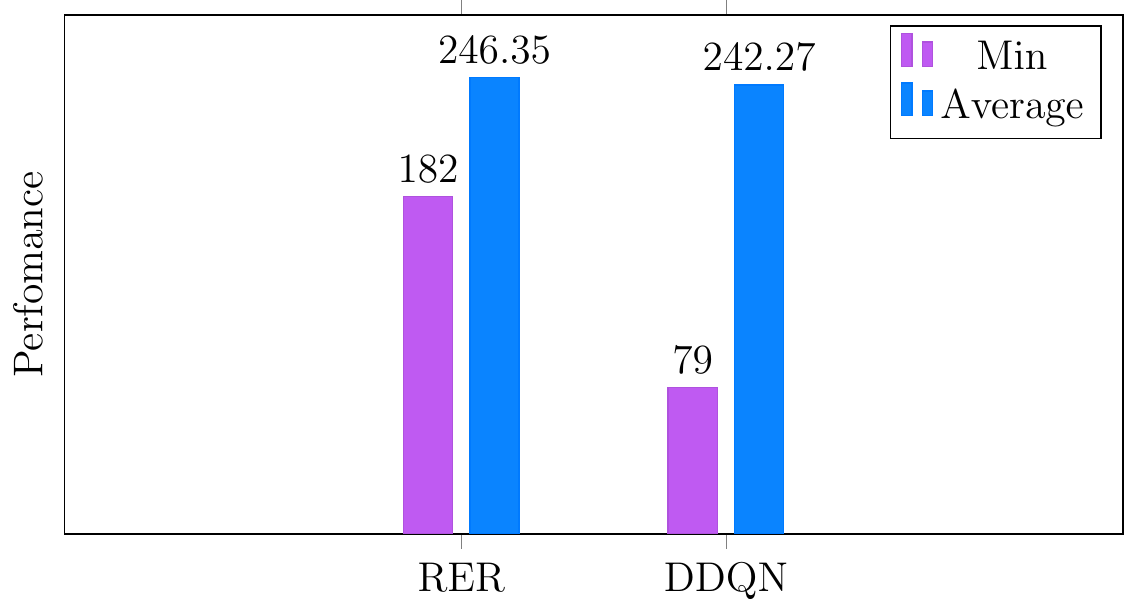}

\caption{Min and average performance of Dueling DQN+RER, Dueling Double DQN+ER algorithms in Breakout.}
\label{perfomanceB}
\end{figure}

40M frames are not enough to well train Neural network with Experience Replay capacity of 100,000 transitions, but RER shows more stable results with these conditions because of better generalization and remembering rarely visited transitions. See Figure \ref{perfomanceB}. Also, it achieves better results on average.

\section{Discussion}
Described approach logically follows from the knowledge about the backup and Prioritized Sweeping method. The presented learning process does not require storing the agent's Neural network's copy while Double DQN does. NN's copy can have a negative impact on learning time and computational resources for it. Both solutions should be supplemented with Experience Replay (either Reverse or a standard one). For these reasons, vanilla DQN with RER is more memory friendly by storing only Q-Network and experience. On the other hand, Reverse Experience Replay's sampling is more complex and takes more time. Because of this, DQN without Target-Network with RER can be used in cases when the complexity of the Neural network's architecture is significantly higher than the complexity of the presented algorithm of sampling, or when it is not possible to store enough experience for standard Experience Replay.

\section{Conclusion}
Reverse order update is a natural method of learning from rewards discovered on rats that accelerates their learning. This method can be adapted for agents with the Neural network to speed up their learning and make the learning process more similar to the way the brain processes the reward. It yields better and more stable learning results from rare transitions when compared to Double DQN on tasks with small memory size and lack of experience. Also, RER shows increased results against DQN with standard Experience Replay. Therefore Reverse Experience Replay can be useful in specific real-world tasks with delayed reward signals.

\section{Funding}
This research did not receive any specific grant from funding agencies in the public, commercial, or not-for-profit sectors.

\section*{Acknowledgements}
I wish to thank Gregory Antonovsky for the feedback and advice that improved the manuscript.

\end{document}